\documentclass[runningheads]{llncs}
\usepackage[T1]{fontenc}
\usepackage{amsmath}
\usepackage{cite}
\makeatletter
\renewcommand{\@biblabel}[1]{#1.}
\makeatother
\usepackage{booktabs}
\usepackage[colorlinks=true, linkcolor=blue, citecolor=blue, urlcolor=blue]{hyperref}

\usepackage{graphicx}
\begin{document}
\title{Building Reasonable Inference for Vision-Language Models in Blind Image Quality Assessment}

\titlerunning{Buliding Reasonable Inference for VLMs in BIQA}
%
%
\author{Yuan Li\inst{1}\textsuperscript{*}\orcidID{0009-0004-2797-0397} \and
Zitang Sun\inst{1}\orcidID{0000-0003-2267-421X} \and
Yen-ju Chen\inst{1}\orcidID{0000-0002-2038-1440} \and
Shin'ya Nishida\inst{1}\orcidID{0000-0002-5098-4752}}
\authorrunning{Li. et al.}
%
\institute{Graduate School of Informatics, Kyoto University,  Kyoto, {606-8501},  {Japan} 
\email{*Corresponding author. E-mail: li.yuan.67n@st.kyoto-u.ac.jp}\\
\email{\{sun.zitang.c09,chen.yen-ju.t05\}@kyoto-u.jp}\\
\email{nishida.shinya.2x@kyoto-u.ac.jp}}
\maketitle              
\begin{abstract}
Blind image quality assessment (BIQA) has recently been significantly improved by leveraging vision-language models (VLMs). Given their semantic reasoning capabilities, it is tempting to assume that these models extract visual features, generate descriptive texts, and perform quality inference in a human-like manner. However, such models sometimes produce textual descriptions of visual attributes that contradict the final quality predictions (e.g., the model outputs ``the image is clear'' but finally assesses the quality as ``poor''). Moreover, the predicted quality often changes unstably during the inference process. These behaviors are obviously inconsistent with human reasoning. To gain insights into the gap between the models and humans, this study investigates what causes contradictory quality assessments and why the assessments are susceptible to change. We first estimate the relationship between the final quality assessments and the generated visual features. Our analysis reveals that the final predictions are, to some extent, not derived through reasoning based on these features, so that the logical correlation between them is relatively weak. Furthermore, by decoding the intermediate layers of the VLM, we observe that the language model often relies only on a limited set of candidate tokens. This behavior further explains why the quality assessments are susceptible to change. To promote human-like logical reasoning in VLMs, we next introduce a two-stage tuning method that explicitly decouples visual perception from quality inference. The VLM is instructed to learn the visual features and conclude the quality solely based on the visual features, respectively, in the two stages. Experimental results on the standard IQA datasets (SPAQ and KONIQ) show that our approach reduces prediction instability from 22.00\% to 12.39\%, and achieves average improvements of 0.3124/0.3507 in terms of SRCC/PLCC across four standard datasets (LIVE, CSIQ, SPAQ, and KONIQ), compared to the baseline. Further analysis and visualization provide evidence that our proposed method not only enhances the model’s stability but also builds a reliable inference.

\keywords{Blind Image Quality Assessment \and Vision-Language Model  \and Stable Reasoning \and Interpretable Inference}
\end{abstract}
\section{Introduction}

\begin{figure}[!t]
  \centering
  \includegraphics[width=\textwidth]{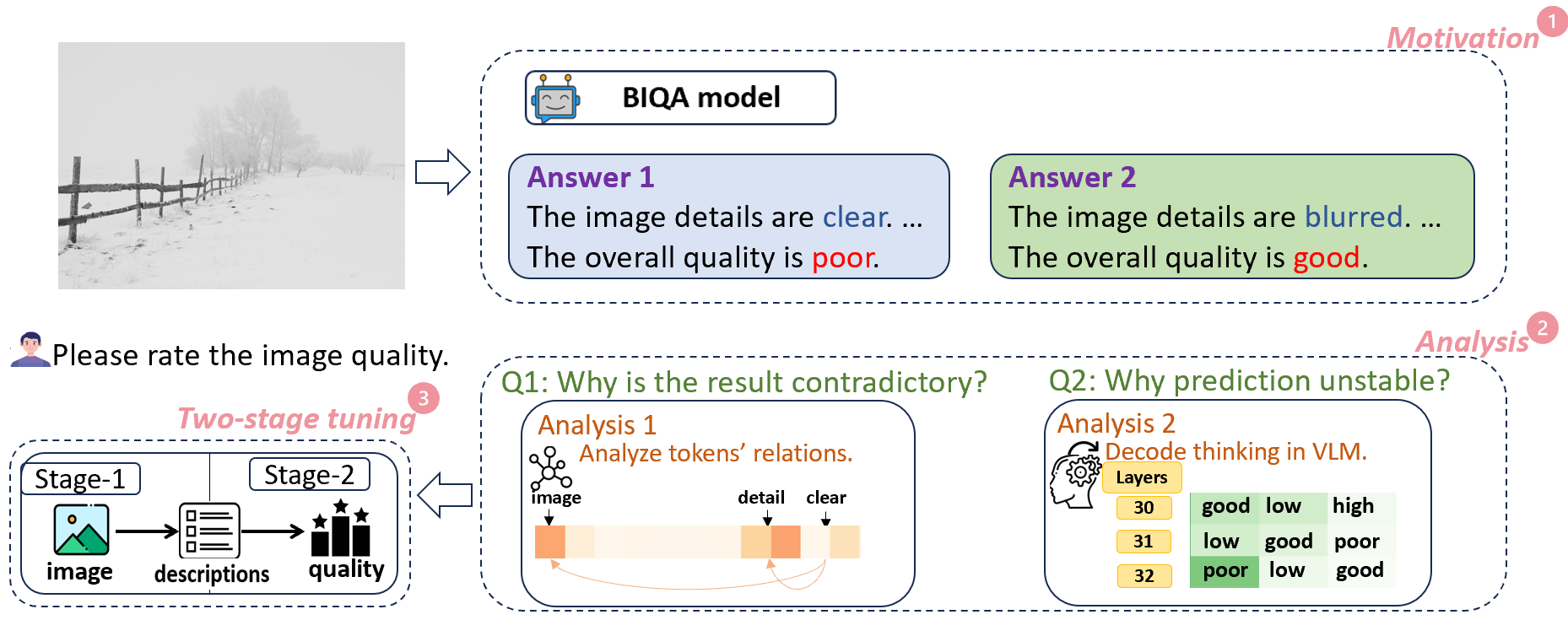}
  \caption{\textbf{Overview.} 
  VLM-based BIQA model \cite{qinstruct} exhibits the contradictory reasoning and generates inconsistent answers across repeated queries. Motivated by this observation, we investigate how the visual features contribute to the final quality predictions. In addition, we inspect the VLM reasoning processing by visualizing information dynamics within the decoder's intermediate layers. Based on our findings, we propose a two-stage tuning method that enables the model to produce more reasonable quality assessments while enhancing the stability of its predictions.
  }
  \label{f1}
\end{figure}

Blind image quality assessment (BIQA) remains a fundamental yet challenging problem in computer vision field, as it requires models to predict image quality in the absence of explicit reference images. Traditional approaches primarily focus on extracting visual features \cite{topiq,musiq} or clustering images based on perceived degradations \cite{arniqa,iqa-semi}. While these methods have achieved moderate success, they often lack a high-level understanding of semantic content and fail to incorporate reasoning mechanisms that mimic human judgment.

Recent advances in vision-language models (VLMs) have opened new possibilities for BIQA by integrating visual perception with language-based reasoning. These models \cite{mplug-owl2,flamingo,llava,mplug2} are typically composed of a vision encoder for extracting low-level visual features, a vision-language projector that transforms these features into a semantic embedding space, and a large language model (LLM) responsible for reasoning jointly over visual and linguistic inputs. Therefore, VLM-based BIQA methods \cite{qinstruct,qalign,qground,seagull} have extended beyond traditional scalar scoring by generating textual descriptions that capture fine-grained image attributes alongside quality predictions. Ideally, such models are expected to follow a human-like reasoning chain: from low-level visual features (e.g., color, composition), to semantic reasoning for quality assessment (e.g., good, poor). 

However, we have observed that VLM-based BIQA models \cite{qinstruct,qground,depictqa}   sometimes exhibit that their quality reasoning is not well aligned with the generated visual features and is highly susceptible to change, as illustrated in Fig.~\ref{f1}. These inconsistencies raise concerns about the reliability of VLM outputs.

To investigate  what causes contradictory quality assessments and  why are these assessments susceptible to change, we firstly estimate the relationship between the quality prediction and basic visual features utilizing the attention connections. The attention visualization indicate that VLMs tend to overfit the final prediction directly to abstract visual embeddings rather than the basic visual features. That somehow explains why the model generates contradictory features and assessments. 

Subsequently, we adopt the logit lens method \cite{backwardlens} to examine the inference process for specific tokens.
Our analysis reveals that VLMs often select a final quality token from multiple competing candidates and that inconsistencies in intermediate layer selections possibly lead to instable predictions.

Motivated by these observations, we are thinking to decouple quality assessment from a direct reliance on abstract visual embeddings and instead encourage the model to focus more explicitly on visual features, thereby promoting greater interpretability and consistency in the overall reasoning process.  Therefore, we propose a two-stage tuning method. In the first stage, we fine-tune the VLM to align image features with fundamental visual descriptions using multi-modal data. In the second stage, the model is trained to infer final quality predictions from these descriptions under a single-modal instruction paradigm. By compelling the model to rely on fundamental visual descriptions rather than directly inferring a score, our two-stage approach significantly improves prediction stability and interpretability. Specifically, on a mixed test set of SPAQ \cite{spaq} and KONIQ \cite{koniq}, it reduces the average instability ratio from 22.00\% to 12.39\% compared to one-stage method \cite{qinstruct} and gain an average 0.3124/0.3507 improvement on SRCC/PLCC metrics across four datasets (SPAQ, KONIQ, LIVE and CSIQ \cite{spaq,koniq,live,csiq}) compared to the baseline model. 

In summary, our main contributions are:
\begin{enumerate}
    \item \textbf{Quality Inference Analysis:} We estimate the relation between quality assessment and basic visual features and identify some existing models bypass visual features which conclude a contradictory quality result. Subsequently, we integrate the logit lens method \cite{backwardlens} into VLM-based BIQA models to elucidate the inference process. By decoding intermediate inference states, to some extent, we explain why the BIQA models generate the unstable predictions.
    \item \textbf{Two-Stage Tuning:} We propose a two-stage framework that decouples visual perception and reasoning. This enforces the model to ground quality predictions on fundamental visual descriptions, thereby mitigating prediction instability and enhancing the logical consistency between descriptive features and the resulting quality assessments.
\end{enumerate}

\section{Related Works}

\subsection{Vision-Language Instruction Tuning}
To effectively process both visual and textual prompts, vision-language models are trained under a multi-modal instruction tuning framework. This paradigm extends instruction tuning from natural language processing (NLP) systems to multi-modal ones. Traditionally, large language models are trained to perform complex real-world tasks under the guidance of language instructions \cite{llm_instruct,llm_self_instruct,t5}, where supervised training relies on conversational interactions between humans and models in a purely textual format.
To integrate LLM capabilities into multi-modal models, it is crucial to establish a universal feature space that enables effective cross-modal information transfer, known as feature alignment. Typically, visual features are projected into the language space \cite{flamingo,llava}, allowing LLMs to process visual inputs without additional assistance. LLaVA \cite{llava} presents a foundational architecture for VLMs, comprising three primary components: a vision encoder, a projector, and a language model. In this work, to ensure fair comparisons with existing BIQA models \cite{qinstruct,qalign}, we adopt mPLUG-Owl2 \cite{mplug-owl2} as our backbone, as illustrated in Fig.~\ref{f2}. This model shares a similar architectural design with LLaVA \cite{llava}, with the key distinction being that the vision projector in mPLUG-Owl2 \cite{mplug-owl2} is substantially more complex than the multi-layer perceptron used in LLaVA \cite{llava}.

\subsection{Blind Image Quality Assessment Using VLMs} 
Recently, the advent of vision-language models has introduced enhanced capabilities for semantic comprehension and interpretable reasoning. Motivated by these advances, researchers \cite{qinstruct,seagull,depictqa} have increasingly integrated VLMs into BIQA frameworks. Notable examples include Q-Instruct \cite{qinstruct}, Q-Ground \cite{qground}, Seagull \cite{seagull} and DepictQA \cite{depictqa}, which have curated extensive datasets and fine-tuned VLMs accordingly. These datasets \cite{qinstruct,qground,seagull} commonly incorporate detailed textual descriptions relating to image quality and local region annotations generated by segmentation techniques (e.g., SAM \cite{sam}), thereby aiming to enhance VLMs’ capability to recognize quality-related visual attributes and generate more interpretable quality assessments.

Although recent VLM-based methods have achieved state-of-the-art performance in image quality prediction \cite{qinstruct,qalign,seagull}, they still occasionally suffer from inconsistent perceptual judgments and exhibit logical contradictions in their reasoning processes. Despite notable empirical improvements, a thorough understanding of the perception and reasoning mechanisms within VLMs remains limited. This raises critical questions about whether VLMs genuinely acquire a generalized understanding of visual quality or merely overfit specific data distributions. To address these issues, in this study, we systematically analyze the internal generation processes of VLMs, aiming to uncover their underlying reasoning patterns. Our findings provide deeper insights into VLM behavior and offer practical guidance for enhancing related vision-language downstream tasks.

\section{Methodology}
\label{method}
\subsection{Preliminary}
\label{3.1}
\begin{figure}
    \centering
    \includegraphics[width=\linewidth]{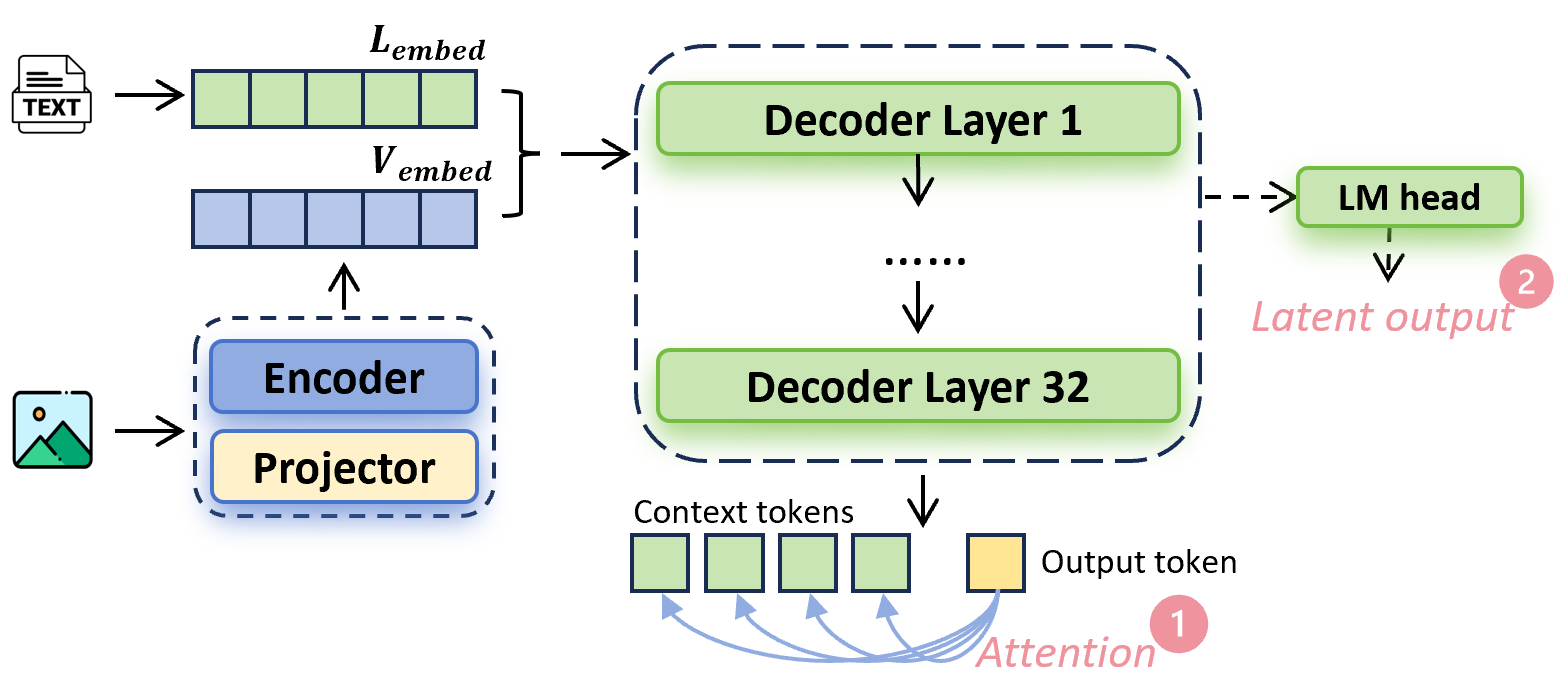}
    \caption{Both textual and visual inputs are projected into a shared semantic embedding space. The language and visual embeddings, denoted as ${L_{embed}, V_{embed}}$, each has 4096 dimensions. The generation of a single token requires processing through 32 decoder layers, where the model sequentially refines its predictions. The latent embeddings can be directly decoded to visualize the intermediate processing as we denoted as latent output. And the attention represents the relation between the current token and its previous context. In BIQA tasks, the context is basic visual features such as degradations or color.}
    \label{f2}
\end{figure}

As illustrated in Fig.~\ref{f2}, prompts and images are first encoded into a shared semantic space. Generally, the text prompt is a template like "Please rate the image quality" to instruct the model to analyze the image quality. Subsequently, LLM processes the embeddings to generate predictions via multiple decoder layers. The VLM-based models will first generate the context tokens, e.g. the composition/color of the image is ..., and the final output of the models is <quality>.

\subsection{Relation Estimation for Quality and Context}
\label{3.2}
During token inference, we investigate how quality predictions are influenced by context tokens. Most modern LLMs, including Llama \cite{llama2} and GPT-4 \cite{gpt4}, adopt a transformer-based architecture. Within this framework, attention values between tokens provide insights into the relationships and contributions of different tokens to the final prediction. In BIQA works, if the model exhibit human-like thinking chains, the prediction token of <quality> should be highly related to the generated context, which are the fundamental visual features. To quantify these influences, we analyze attention values, which measure the impact of context tokens—including both multi-modal input tokens and previously generated tokens—on the current token’s prediction. The inference connections are formally represented by the attention mechanism, computed as follows:
\begin{align} 
{Attention(i,j)} = \text{Softmax} \left(\frac{q_{i}\cdot k_{j}}{\sqrt{d}}\right) v_{i} \label{eq2} 
\end{align}
where ${i,j}$ denote the $i$-th and $j$-th tokens, while ${q, k, v}$ correspond to their respective query, key, and value vectors. This formulation quantify the influence of image tokens on the language model’s inference process, offering deeper insights into the reasoning mechanisms within VLMs. By visualizing this reasoning process, we find that <quality> prediction tends to ignore the context tokens but straightly fit to some unexplainable visual embeddings, as shown in Fig.~\ref{f5}.

\subsection{How VLMs Infer in the Middle Layers}
\label{3.3}
\begin{figure}[h]
    \centering
    \includegraphics[width=\linewidth]{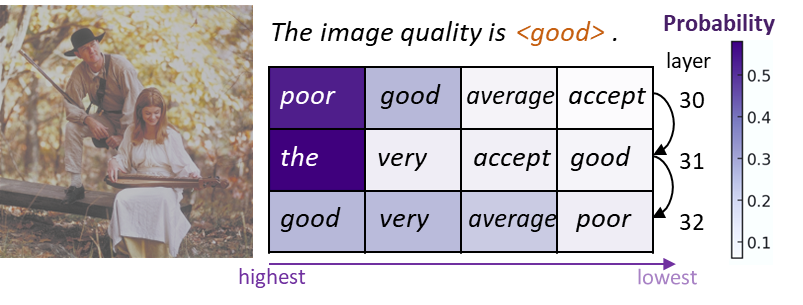}
    \caption{\textbf{Latent Output.} Decoding the latent from intermediate layers of a BIQA model \cite{qinstruct}, four most probable candidates of output token are visualized. In 30-th layer, the most probable token is ``poor'' while the output token is ``good''. It is reasonable in LLM system but does not fit the human reasoning. This image is from the CSIQ dataset \cite{csiq} and has a quality score of 0.416 (where 1.0 represents the highest quality). }
   
    \label{f3}
\end{figure}
The inference capability of VLMs heavily depends on the underlying language model. In the NLP community, researchers have employed projection techniques to map weights and hidden states onto the language model’s vocabulary, providing insights into the information flow within its internal layers \cite{backwardlens}. Inspired by this approach, we introduce a similar methodology to analyze the learning and inference mechanisms in our work.
To examine latent outputs, we utilize the language model’s head layer to predict the most probable next tokens based on the current state. Our analysis reveals that intermediate outputs often differ from or even contradict the final predictions, as illustrated in Fig.~\ref{f3}. Specifically, we observe a transition in quality predictions—from ``poor'' at the 30-th layer to ``good'' at the output layer. Notably, when the same input is tested again, the final prediction may switch back to ``poor'', highlighting the inconsistency in inference. In Section \ref{4.3.1}, we systematically discuss the evolution of prediction tokens. Additionally, we observe that semantically meaningful information tends to emerge in the later layers. Therefore, we decode the outputs starting from the 30-th of 32 decoder layers, tracking changes in token probability distributions to gain deeper insights into the instability of predictions.

\subsection{Two-Stage Tuning}
\label{3.4}
\paragraph{\textbf{Data Preparation}}
Most multi-modal BIQA datasets \cite{qinstruct,qground,seagull} consist of original images paired with corresponding quality-related descriptions. They are using one-stage tuning method to conclude the image quality. To enhance the robustness of the inference stage, we decompose the Q-Pathway \cite{qinstruct} dataset into two distinct stages.
\begin{figure}[!h]
    \centering
    \includegraphics[width=1\linewidth]{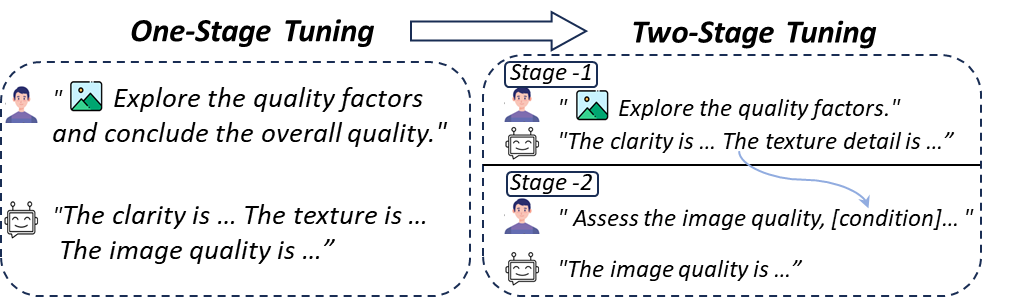}
     \caption{From One-Stage to Two-Stage Tuning.}
    \label{f4}
    \vspace{-5mm}
\end{figure}
In the first stage, the VLM learns fundamental visual descriptions using visible image tokens. In the second stage, the model is solely provided with language prompts, enabling it to assess image quality based on the given descriptions.

\paragraph{\textbf{Fine-tuning}}
The baseline model \cite{mplug-owl2} is pre-trained on multi-modal instruction data, which enables it to achieve a strong alignment between images and text. During the fine-tuning stage, we apply the label smoothing negative log-likelihood loss function, as defined in Eq.~\ref{eq2}, to the output of the language model.

\begin{align} \small \mathcal{L}(\boldsymbol{\theta})
= (1 - \epsilon)  \cdot \Bigl[-\log p_\theta\bigl(y \mid x\bigr)\Bigr] + \epsilon  \cdot \frac{1}{C} \Bigl[\sum_{c=1}^{C} \bigl(-\log p_\theta(c \mid x)\bigr)\Bigr] \label{eq2} \end{align}
Here, $\theta$ represents the model parameters, $\{x, y\}$ denote the input and true label, $C$ is the number of classes, $\epsilon$ is a hyper-parameter controlling the smoothing factor, and $p$ is the conditional probability.

\section{Experiments}
\label{experiments}
\subsection{Implementation Settings}
\label{4.1}
We adopt a pre-trained mPLUG-Owl2-7b checkpoint \cite{mplug-owl2} as the backbone model. For fine-tuning, we randomly select 6,720 samples from the Q-Pathway \cite{qinstruct} dataset, with 6,000 samples allocated to the training set. The fine-tuning process is conducted on four RTX A6000 GPUs.
We employ a two-stage fine-tuning approach, where stage one is trained for 6,000 iterations, followed by stage two for another 3,000 iterations. In contrast, the baseline model \cite{mplug-owl2} is fine-tuned using a one-stage tuning strategy for 9,000 iterations. All fine-tuning experiments utilize the AdamW optimizer \cite{adamw}, with $\beta_1 = 0.9$, $\beta_2 = 0.98$, and $\epsilon = 1e-6$.

\subsection{Instability \& Quality Evaluation}
\label{4.2}
\begin{table}[h]
    \caption{Instability Ratio on IQA Datasets}
    \centering
    \begin{tabular}{c|c|c}
        \toprule
        \textbf{Instability }(\%) $\downarrow
         $ & One-Stage \cite{qinstruct} & Two-Stage \\
        \hline
          {\small SPAQ \cite{spaq}+KONIQ \cite{koniq}} & 22.00 ($\pm$0.08) & \textbf{12.39} ($\pm$0.52)  \\
          \hline
         LIVE \cite{live} &16.27  ($\pm$0.47) & \textbf{4.09} ($\pm$0.11) \\
         \hline
         CSIQ \cite{csiq} & 13.17 ($\pm$0.27) & \textbf{6.08} ($\pm$0.63) \\
        \bottomrule
    \end{tabular}
    \label{t1}
\end{table}
We evaluate prediction stability on both authentic and synthetic samples. The authentic samples are randomly selected from the SPAQ \cite{spaq} and KONIQ \cite{koniq} datasets, forming a mixed set of 3,000 samples. Meanwhile, the synthetic samples are from the LIVE \cite{live} and CSIQ \cite{csiq} datasets, comprising 20\% of each dataset.
The evaluation is conducted across three sessions. In each session, quality predictions are tested five times per sample to assess stability. To quantify prediction stability, we define the instability ratio as the frequency of generating inconsistent or uncommon predictions, where a lower ratio indicates greater stability. Our fine-tuned method significantly reduces the instability ratio, demonstrating that the proposed approach effectively enhances the robustness and reliability of VLM-based quality predictions.
\begin{table}[!h]
    \caption{\textbf{Quality Results} of Spearman's correlation coefficient (SRCC $\uparrow$) and Pearson correlation coefficient (PLCC $\uparrow$). Higher is better for both metrics. Q-Instruct model \cite{qinstruct} result is reproduced by training and testing under the same condition. Q-Align \cite{qalign} model is tested using their pretrained checkpoint which included testing samples and trained on more than ten times larger data.}
    \centering
    \begin{tabular}{c|c|c|c|c}
        \toprule
        \textbf{Datasets} 
          & baseline \cite{mplug-owl2}&Q-Instruct \cite{qinstruct} & Q-Align \cite{qalign}& Ours \\
        \hline
          {\small SPAQ \cite{spaq}} & 0.1364/0.1267 & 0.3229/0.3087& \textbf{0.8381}/0.7341 &0.7437/\textbf{0.7935} \\
         \hline
          {KONIQ \cite{koniq}} & 0.0262/0.0419& 0.1191/0.1153& \textbf{0.8340/0.7086}&0.5854/0.6460  \\
          \hline
         LIVE \cite{live} & 0.0475/0.0537&0.4938/0.5173&\textbf{0.89845/0.8180}&0.7006/0.7161 \\
         \hline
         CSIQ \cite{csiq} & -0.0205/-0.0298 &0.3641/0.3586&\textbf{0.7419/0.6348}&0.5201/0.6176\\
         \hline
         Inference time ($\downarrow$)&0.1605s&\textbf{0.1467s}&0.1603s&0.2044s \\
        \bottomrule
    \end{tabular}
    \vspace{-3mm}
    \label{t2}
\end{table}

Limited to the training data, our model only exhibits the comparable performance, compared to the advanced models. But under the fair comparison, our model still exhibits remarkable performance with average improvements of 0.3124/0.3507 in terms of SRCC/PLCC metrics, compared to the one-stage model \cite{qinstruct}, with a slight trade-off on inference time.

\subsection{Intermediate Features Analysis}
\begin{figure}[h]
  \centering
  \includegraphics[width=\textwidth]{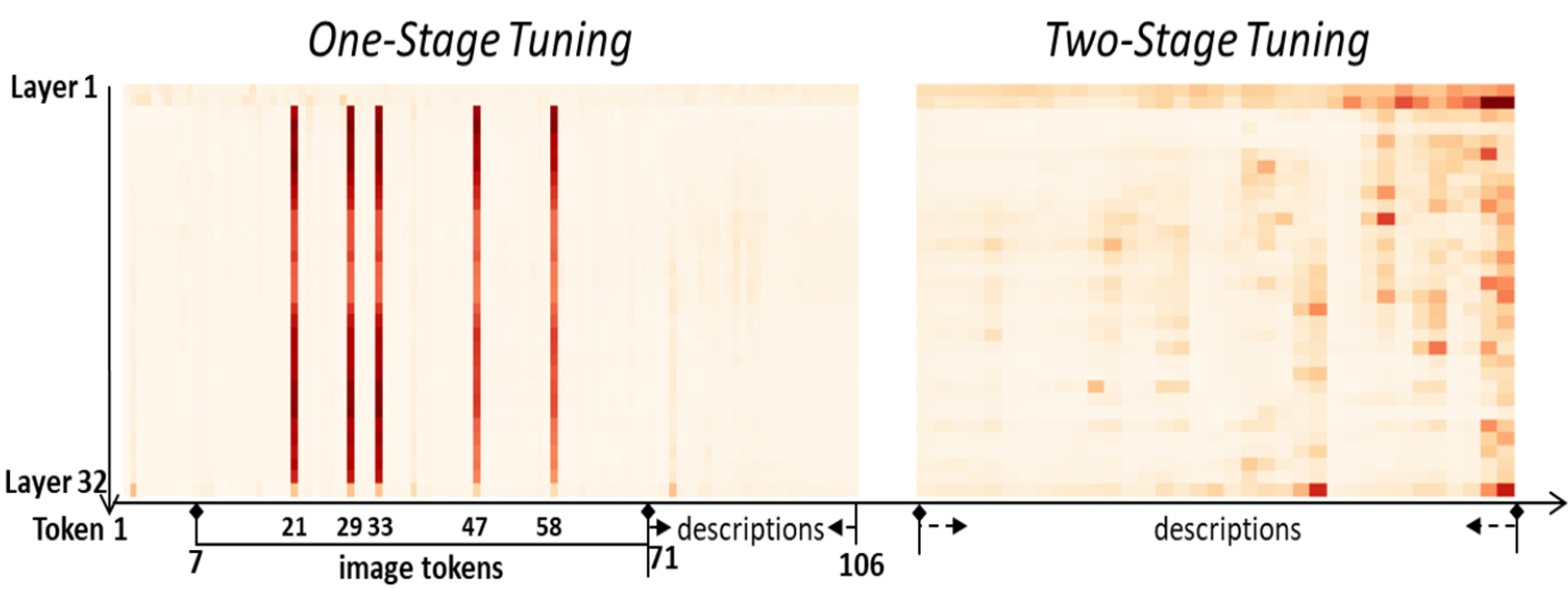}
  \caption{ \textbf{Inference Visualization.} This exhibits the inference of the $\langle quality\rangle$ token. The attention maps are computed and averaged across 720 samples, as described in Section \ref{4.3.1}. In the left panel, we present the one-stage tuning model \cite{qinstruct,mplug-owl2}, where the input prompts include 65 image tokens. From token 71 onward, the generated tokens primarily describe the visual descriptions of the image. The right panel shows the inference process of the two-stage tuning model, where all tokens correspond to image quality descriptions. Compared to the one-stage tuning model \cite{qinstruct,mplug-owl2}, which focuses on a limited set of visual embedding tokens, the two-stage tuning model exhibits a broader attention distribution across important description-related tokens.}
  \label{f5}
\end{figure}

\subsubsection{Inference Attentions}
\label{4.3.1}
Fig.~\ref{f5} visualizes the inference process when the models predict image quality, with attention maps revealing key patterns in the reasoning process.
A closer examination provides unexpected insights into the VLM inference process. In the one-stage tuning model \cite{qinstruct, mplug-owl2}, the prediction stages consistently exhibit strong connections to five specific image embedding tokens. This suggests that the model primarily bases its quality predictions on these unexplainable tokens, which bypass the generated visual descriptions and finally lead to inconsistent outputs. Comparing to the one-stage model, our proposed two-stage tuning methodology enforces quality inference based on fundamental descriptions rather than isolated visual tokens. Given that the effective tokens are distributed separately within the context, the attention maps are expected to be associated with dynamic, unfixed tokens. The results confirm this hypothesis, further explaining how two-stage tuning establishes a more robust and structured quality reasoning process compared to the one-stage model \cite{qinstruct, mplug-owl2}. Additionally, this finding highlights a potential future direction, where image quality predictions could be directly derived from these key image embeddings, offering a more efficient and stable approach for BIQA models.

\subsubsection{Latent Output}
\label{4.3.2}
\begin{figure}[h]
    \centering
    \includegraphics[width=\textwidth]{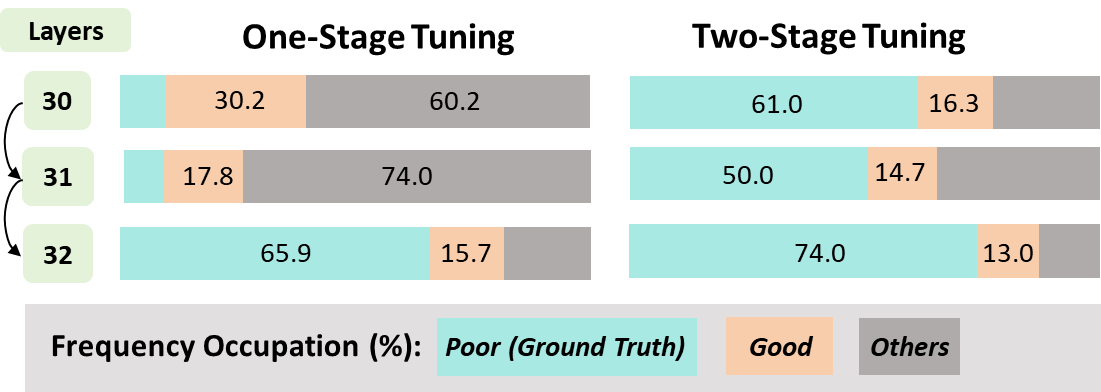}
    \caption{\textbf{Latent Output after 30-th Layers.} Evaluate our approach on 300 samples from the Q-Pathway test dataset \cite{qinstruct}, decoding the latent outputs from intermediate layers. Extract samples classified as having ``poor'' quality and visualize the corresponding inference chains. Two-Stage Tuning built a more smooth prediction. }
    \label{f6}
\end{figure}
We evaluate 720 samples from the Q-Pathway dataset \cite{qinstruct} across three testing sessions. Our two-stage tuning model improves the overall accuracy from 52.50 ($\pm$0.37)\% to 61.20 ($\pm$0.49)\% . To further investigate the model's inference process, we select 300 samples classified as ``poor'' quality and compute the token frequency distribution across intermediate layers, as illustrated in Fig.~\ref{f6}. In earlier layers, the one-stage tuning model \cite{qinstruct,mplug-owl2} exhibits a high probability of predicting ``good'' quality, potentially leading to instability in final predictions. Comparing to the one-stage model, which demonstrates a sudden probability increase in the final layer, our two-stage model maintains a more stable probability distribution, allowing relevant tokens to emerge earlier in intermediate layers. Our approach achieves an 8.7\% improvement in prediction accuracy on the test samples. This enhanced token stability not only contributes to a more robust and reliable image quality prediction process but also underscores the effectiveness of the proposed two-stage tuning strategy in improving inference consistency.

\section{Discussion \& Limitation}
\label{dl}
\subsection{The Gap between VLMs and Humans in BIQA}
Previously, researchers \cite{qinstruct,depictqa} stated that quality predictions could be inferred from perceived visual features, rendering VLM-based models both explainable and human-aligned. However, our findings reveal that current VLMs often bypass this intended reasoning pathway and instead directly regress to quality scores. In doing so, the model partially degenerates into a shallow regression system, thereby undermining the reasoning potential embedded in its LLM component. This shortcut behavior introduces not only logical gaps between visual descriptions and predictions, but also increases the instability of generated outputs across inference instances.

Moreover, many existing multi-modal training datasets for BIQA tasks are constructed using similar templates. Intuitively, such templated formats may lead the VLM to degenerate into a mere classification model, or be treated as meaningless biases due to their overly repetitive patterns. To achieve more human-like quality assessment, it may be necessary to shift our focus toward modeling the underlying cognitive mechanisms that drive human judgment, rather than relying solely on surface-level supervision. Estimating and incorporating these deeper behavioral patterns could help guide the model to reason in a more interpretable and generalizable manner.

Furthermore, decoupling the perception and reasoning processes in multi-modal models offers advantages for the development of explainable systems. This separation enables more precise analysis of specific model capabilities, such as assessing whether the model exhibits limited visual perception or whether it correctly interprets human reasoning. However, this decoupling also introduces a critical question: if an error occurs during the perception stage, should it be propagated to the reasoning process? We argue that it should. Even humans may misinterpret visual inputs, which can lead to biased judgments. Nevertheless, the reasoning process can still maintain logical consistency, despite originating from an incorrect perception.

Additionally, prior non-VLM approaches \cite{reiqa,arniqa} have primarily focused on contrastive learning combined with linear regression, and have demonstrated strong performance in quality prediction tasks. However, these methods generally lack interpretability when applied to real-world image scenes. While the introduction of VLMs does increase inference costs, they offer a promising direction for integrating human priors and facilitating future research in explainable BIQA. Nonetheless, we acknowledge that reducing model complexity is crucial. Large-scale VLM architectures may contain redundant parameters that are unnecessary for the specific demands of BIQA. Therefore, model distillation could be a promising strategy to improve both efficiency and scalability.

\subsection{Limitation}
This study has two main limitations. First, although our model emphasizes interpretability through a structured design, its performance in terms of SRCC and PLCC metrics does not reach state-of-the-art levels, likely due to constraints imposed by the scale and diversity of the training data. Since our focus is on explainable systems, this study excludes comparisons with non-VLM models that do not provide interpretable outputs. Second, our experimental evaluation is restricted to a limited set of VLM architectures. In future work, we plan to extend our analysis to a broader range of VLM-based BIQA frameworks to better assess the generalizability and robustness of our proposed approach.

\section{Conclusion}
In this work, we investigate the underlying sources of logical inconsistencies and prediction instability in current BIQA models. By estimating the relationship between quality predictions and fundamental visual features, we identify that some contradictory results arise from bypassing the intended reasoning process. Further analysis of the intermediate layers reveals how unstable predictions are generated. Motivated by these insights, we propose a two-stage tuning strategy for BIQA tasks that guides the VLM to reason in a more human-like manner. Experimental results demonstrate that our method improves both the logical coherence and stability of the model's predictions. We hope our findings will inspire further research on interpretable and human-aligned BIQA approaches.

\begin{credits}
\subsubsection{\ackname} This work was supported by Spring Fellowship (Grant Number JPMJFS2123); and KAKENHI (Grant Number 24H00721).

\subsubsection{\discintname}
The authors have no competing interests to declare that are
relevant to the content of this article.
\end{credits}
%
%
%

\begin{thebibliography}{8}
\bibitem{qinstruct}
Wu, Haoning, et al. "Q-instruct: Improving low-level visual abilities for multi-modality foundation models." Proceedings of the IEEE/CVF conference on computer vision and pattern recognition. 2024.


\bibitem{topiq}
Chen, Chaofeng, et al. "Topiq: A top-down approach from semantics to distortions for image quality assessment." IEEE Transactions on Image Processing (2024).

\bibitem{musiq}
Ke, Junjie, et al. "Musiq: Multi-scale image quality transformer." Proceedings of the IEEE/CVF international conference on computer vision. 2021.

\bibitem{arniqa}
Agnolucci, Lorenzo, et al. "Arniqa: Learning distortion manifold for image quality assessment." Proceedings of the IEEE/CVF Winter Conference on Applications of Computer Vision. 2024.

\bibitem{iqa-semi}
Prabhakaran, Vishnu, and Gokul Swamy. "Image quality assessment using semi-supervised representation learning." Proceedings of the IEEE/CVF Winter Conference on Applications of Computer Vision. 2023.

\bibitem{mplug-owl2}
Ye, Qinghao, et al. "mplug-owl2: Revolutionizing multi-modal large language model with modality collaboration." Proceedings of the ieee/cvf conference on computer vision and pattern recognition. 2024.

\bibitem{flamingo}
Alayrac, Jean-Baptiste, et al. "Flamingo: a visual language model for few-shot learning." Advances in neural information processing systems 35 (2022): 23716-23736.

\bibitem{llava}
Liu, Haotian, et al. "Visual instruction tuning." Advances in neural information processing systems 36 (2023): 34892-34916.

\bibitem{mplug2}
Xu, Haiyang, et al. "mplug-2: A modularized multi-modal foundation model across text, image and video." International Conference on Machine Learning. PMLR, 2023.

\bibitem{qalign}
Wu, Haoning, et al. "Q-ALIGN: teaching LMMs for visual scoring via discrete text-defined levels." In Proceedings of the 41st International Conference on Machine Learning (ICML'24). 2024.

\bibitem{qground}
Chen, Chaofeng, et al. "Q-ground: Image quality grounding with large multi-modality models." Proceedings of the 32nd ACM International Conference on Multimedia. 2024.

\bibitem{seagull}
Chen, Zewen, et al. "SEAGULL: No-reference Image Quality Assessment for Regions of Interest via Vision-Language Instruction Tuning." arXiv preprint arXiv:2411.10161 (2024).

\bibitem{depictqa}
You, Zhiyuan, et al. "Depicting beyond scores: Advancing image quality assessment through multi-modal language models." European Conference on Computer Vision. Cham: Springer Nature Switzerland, 2024.

\bibitem{backwardlens}
Shahar Katz, Yonatan Belinkov, Mor Geva, and Lior Wolf. "Backward Lens: Projecting Language Model Gradients into the Vocabulary Space." In Proceedings of the 2024 Conference on Empirical Methods in Natural Language Processing, pages 2390–2422, Miami, Florida, USA. Association for Computational Linguistics. 2024.

\bibitem{spaq}
Fang, Yuming, et al. "Perceptual quality assessment of smartphone photography." Proceedings of the IEEE/CVF conference on computer vision and pattern recognition. 2020.

\bibitem{koniq}
Hosu, Vlad, et al. "KonIQ-10k: An ecologically valid database for deep learning of blind image quality assessment." IEEE Transactions on Image Processing 29 (2020): 4041-4056.

\bibitem{live}
Ghadiyaram, Deepti, and Alan C. Bovik. "Massive online crowdsourced study of subjective and objective picture quality." IEEE Transactions on Image Processing 25.1 (2015): 372-387.

\bibitem{csiq}
Larson, Eric C., and Damon M. Chandler. "Most apparent distortion: full-reference image quality assessment and the role of strategy." Journal of electronic imaging 19.1 (2010): 011006-011006.

\bibitem{llm_instruct}
Ouyang, Long, et al. "Training language models to follow instructions with human feedback." Advances in neural information processing systems 35 (2022): 27730-27744.


\bibitem{llm_self_instruct}
Wang, Yizhong, et al. "Self-instruct: Aligning language models with self-generated instructions." arXiv preprint arXiv:2212.10560 (2022).

\bibitem{t5}
Chung, Hyung Won, et al. "Scaling instruction-finetuned language models." Journal of Machine Learning Research 25.70 (2024): 1-53.

\bibitem{sam}
Kirillov, Alexander, et al. "Segment anything." Proceedings of the IEEE/CVF international conference on computer vision. 2023.

\bibitem{llama2}
Touvron, Hugo, et al. "Llama 2: Open foundation and fine-tuned chat models." arXiv preprint arXiv:2307.09288 (2023).

\bibitem{gpt4}
Achiam, Josh, et al. "Gpt-4 technical report." arXiv preprint arXiv:2303.08774 (2023).

\bibitem{adamw}
Loshchilov, Ilya, and Frank Hutter. "Fixing weight decay regularization in adam." arXiv preprint arXiv:1711.05101 5 (2017): 5.

\bibitem{reiqa}
Saha, Avinab, Sandeep Mishra, and Alan C. Bovik. "Re-iqa: Unsupervised learning for image quality assessment in the wild." Proceedings of the IEEE/CVF conference on computer vision and pattern recognition. 2023.

\bibitem{arniqa}
Agnolucci, Lorenzo, et al. "Arniqa: Learning distortion manifold for image quality assessment." Proceedings of the IEEE/CVF Winter Conference on Applications of Computer Vision. 2024.


\end{thebibliography}
%

\end{document}